\documentclass[letterpaper, 10 pt, twocolumn]{article}

\usepackage[utf8]{inputenc}
\usepackage{graphicx}
\usepackage[font=small,skip=0pt]{caption}
\usepackage{color, soul}
\usepackage{mathrsfs}
\usepackage{tabularx}
\usepackage{amsmath} 
\usepackage{amssymb}  
\usepackage{amsfonts}
\usepackage{booktabs}
\usepackage{mathtools}
\usepackage{hyperref}
\usepackage{subcaption}
\usepackage{array}
\usepackage{adjustbox}
\usepackage{multirow}
\usepackage{floatrow}

\newfloatcommand{capbtabbox}{table}[][\FBwidth]

\newcommand{\starnote}[1]{}

\usepackage[compact]{titlesec}
\titlespacing{\section}{0pt}{2ex}{1ex}
\titlespacing{\subsection}{0pt}{1ex}{0ex}
\titlespacing{\subsubsection}{0pt}{0.5ex}{0ex}
\newcolumntype{L}{>{\arraybackslash}m}
\newcommand{\eg}{\emph{e.g.}, }
\newcommand{\ie}{\emph{i.e.}, } 
\newcommand{\etal}{\textit{et al}.~}

\title{\LARGE \bf
Autonomous robotic re-alignment for \\face-to-face underwater human-robot interaction\thanks{This work was supported in part by the Science, Mathematics, and Research for Transformation (SMART) Scholarship provided through the US Department of Defense.}
}

\author{Demetrious T. Kutzke$^{1,\dagger}$, Ashwin Wariar$^{2,\dagger}$, and Junaed Sattar$^{3,}$%
\thanks{The authors are with the Department of Computer Science \& Engineering and the Minnesota Robotics Institute,
        University of Minnesota--Twin Cities, Minneapolis, MN 55455, USA {\tt\small \{$^{1}$kutzk015, $^{2}$waria012, $^{3}$junaed\}@umn.edu}}%
}
\date{November 26, 2023}

\begin{document}

\maketitle
\thispagestyle{empty}
\pagestyle{empty}



\makeatletter
\DeclareRobustCommand{\change}{%
  \@bsphack
  \normalcolor 
  \@esphack
}
\DeclareRobustCommand{\stopchange}{%
  \@bsphack
  \normalcolor
  \@esphack
}
\makeatother

\begin{abstract}
The use of autonomous underwater vehicles (AUVs) to accomplish traditionally challenging and dangerous tasks has proliferated thanks to advances in sensing, navigation, manipulation, and on-board computing technologies. Utilizing AUVs in underwater human-robot interaction (UHRI) has witnessed comparatively smaller levels of growth due to limitations in bi-directional communication and significant technical hurdles to bridge the gap between analogies with terrestrial interaction strategies and those that are possible in the underwater domain. A necessary component to support UHRI is establishing a system for safe robotic-diver approach \change to establish face-to-face communication \stopchange that considers non-standard human body pose. In this work, we introduce a stereo vision system for enhancing UHRI that utilizes three-dimensional reconstruction from stereo image pairs and machine learning for localizing human joint estimates. We then establish a convention for a coordinate system that encodes the direction the human is facing with respect to the camera coordinate frame. This allows automatic setpoint computation that preserves human body scale and can be used as input to an image-based visual servo control scheme. We show that our setpoint computations tend to agree both quantitatively and qualitatively with experimental setpoint baselines. The methodology introduced shows promise for enhancing UHRI by improving robotic perception of human orientation underwater.
\end{abstract}
\section{Introduction}

\begin{figure}
\vspace{3mm}
\includegraphics[width=1.0\columnwidth]{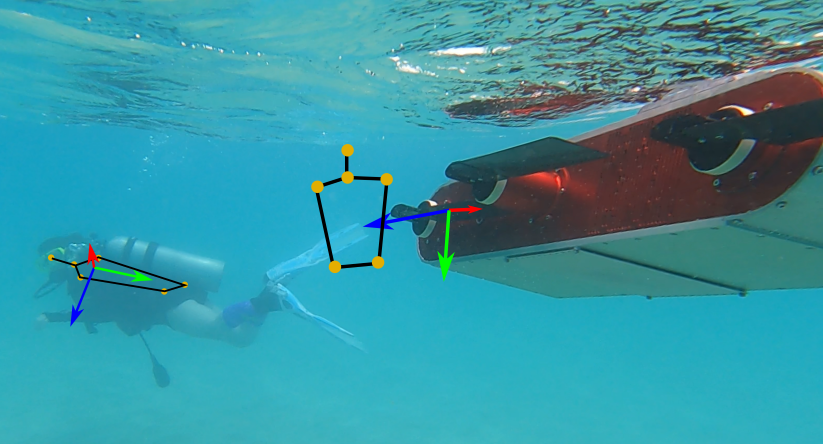}
\caption{Underwater human robot interaction is enhanced by the robot's ability to re-orient itself with respect to the diver, rather than requiring the diver to re-orient with respect to the robot.}
\label{fig:aqua_Chelsey}
\end{figure}

Problems associated with humans and robots interacting, also referred to as human-robot interaction (HRI), is well-studied in controlled terrestrial environments \cite{mataric2018relevance}.
Innovations in HRI have bolstered adoption of these technologies into many areas of life, such as manufacturing \cite{cherubini2016collaborative}, medicine \cite{su2018safety}, long-term care of the elderly \cite{fasola2013socially}, military applications \cite{jentsch2016human}, and the underwater domain \cite{fulton2019robot}. This is due in part to the benefit of allowing robots to take on the dirty, dull, and dangerous tasks \cite{takayama2008beyond} that would otherwise place humans in direct harm or assist in situations in which it is not possible for the human to provide the level of persistent attention required, as in the case of long-term care facilities. The thought goes that off-loading these tasks to robots will allow humans to interact with them from relative safety or convenience, while also performing oversight \cite{sheridan2016human}.

Underwater human-robot interaction (UHRI) is much more challenging, because the underwater domain presents a formidable environment for robotic sensing. It lacks many of the benefits of the terrestrial domain such as high bandwidth radio communication, \eg wireless internet and Bluetooth, consistent lighting conditions, and localization via global-positioning satellites. Underwater robots must utilize alternative methods to perform the same tasks as their terrestrial counterparts.
Often, expensive sonar-based techniques are used to perform navigation and localization \cite{corke2007experiments}. Utilizing visual sensors is also challenging, because differences in salinity and particulates in the water can occlude and distort imagery. Creative techniques must be employed to enhance vision underwater when the conditions are especially degraded \cite{edge2020generative, islam2020, islam2020RSS}. However, even with the challenges of visual imagery, there are instances where utilizing camera data is preferable to expensive and invasive acoustic systems. For example, when acoustically susceptible marine life such as dolphins or whales are present \cite{hastings2008coming}, in these instances, utilizing visual sensing, which is less invasive, is both ethical and beneficial to the preservation of the marine life. 
Divers operate in a similarly sensory-deprived state. Scuba masks occlude peripheral vision or at the very least can reduce the diver's ability to see or perceive dynamic robotic gestures underwater; acoustic signals are degraded by inhalation and exhalation through the breathing regulator, which significantly reduces the diver's ability to hear; and environmental conditions such as strong currents, silting from sediment in the water column, and frigid water temperatures all contribute to generally high cognitive loads. Sensory-deprived states for both robots and humans means that in complex UHRI scenarios, where communication is critical from both robot-to-human (R2H) and human-to-robot (H2R), there is a high-probability of information loss. We argue that because of these conditions, the robot must have the ability to autonomously establish face-to-face (F2F) communication. \textit{F2F} communication reduces the probability of information loss by ensuring that the robot and the human are within a safe distance and in full view of each other. The diver can see the robot's movements and vice-versa. The diver also has the best chance of hearing any acoustically communicated information. 
 
To achieve this F2F configuration, we propose a stereo vision algorithm to autonomously compute a desired feature setpoint, which can be used for visual servo control schemes. This eliminates the need for human-engineered features and equips the robot with the ability to infer a desired F2F setpoint from nonstandard body poses that preserves scale. Scale is important to ensure safe and consistent approach distances for divers of different shapes. To our knowledge, the problem of autonomously establishing F2F communication \change underwater \stopchange has not been considered for general body poses or instances where the robot is not already within a safe communication distance and can perceive the human diver's face.

\begin{figure}[t]
\vspace{1.75mm}
\captionsetup[subfigure]{labelformat=empty}
    \centering
    \begin{subfigure}[t]{0.24\columnwidth}
        \includegraphics[width=1.0\textwidth]{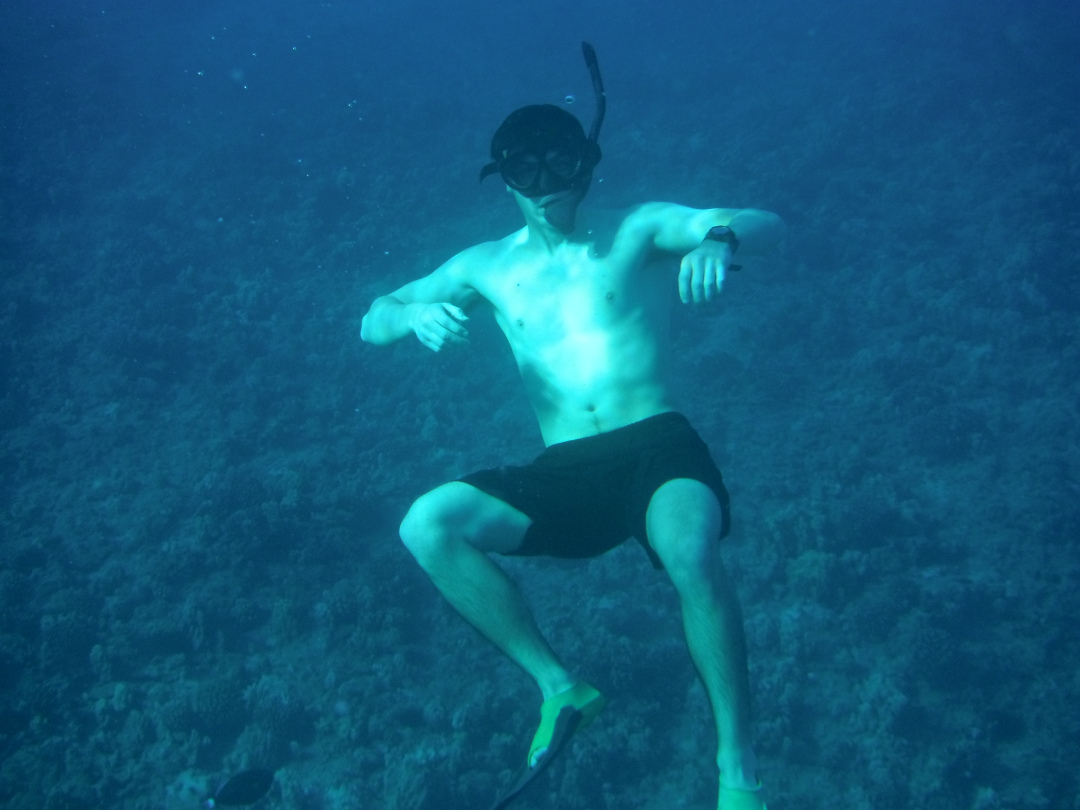}
        \caption{}
    \end{subfigure}
    \begin{subfigure}[t]{0.24\columnwidth}
        \includegraphics[width=1.0\textwidth]{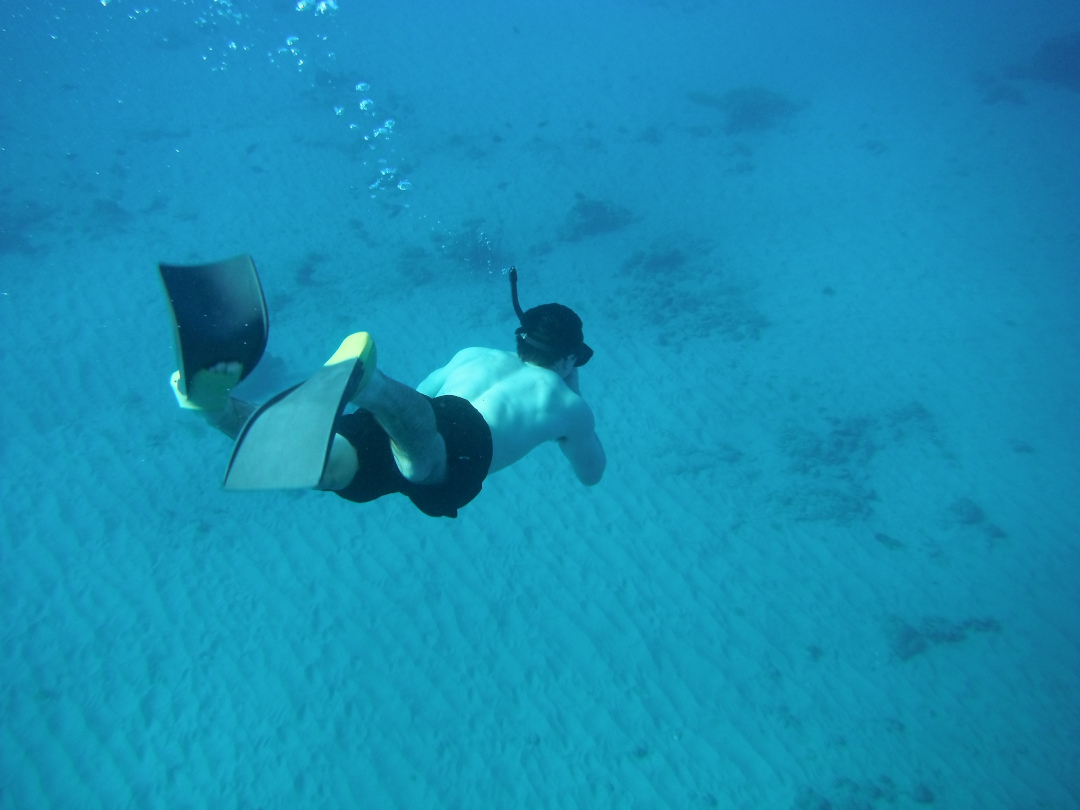}   
        \caption{}        
    \end{subfigure}  
    \begin{subfigure}[t]{0.24\columnwidth}
        \includegraphics[width=1.0\textwidth]{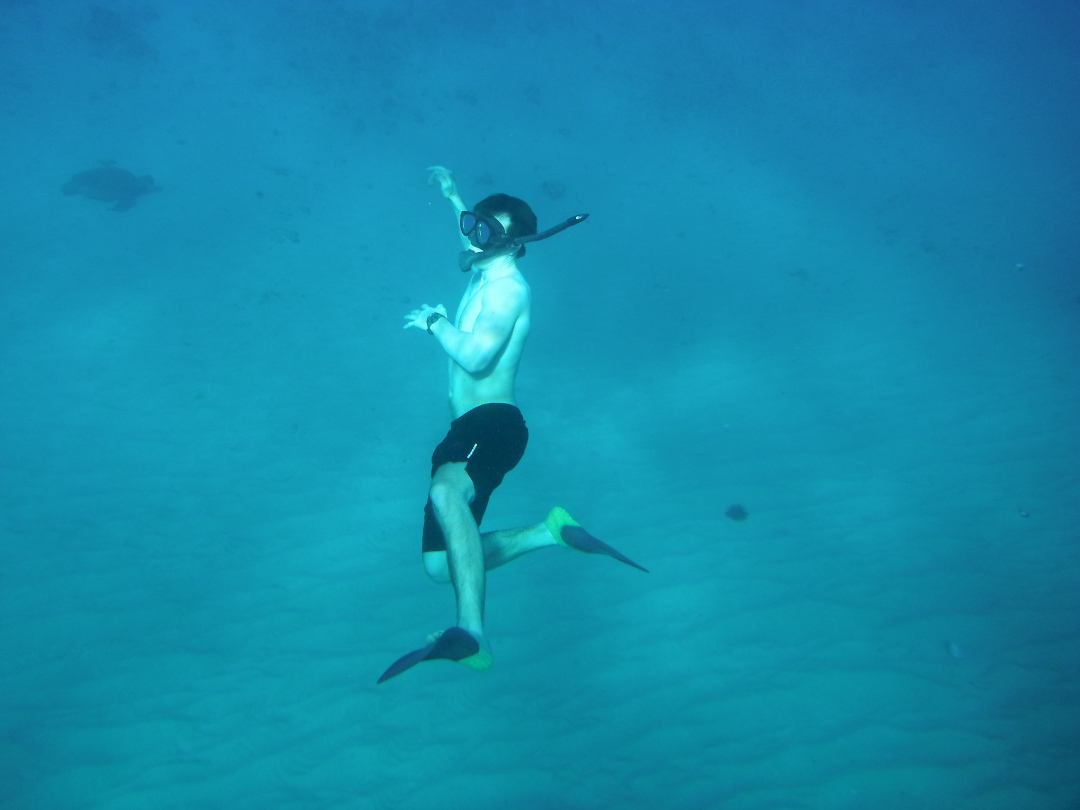}   
        \caption{}        
    \end{subfigure}
    \begin{subfigure}[t]{0.24\columnwidth}
        \includegraphics[width=1.0\textwidth]{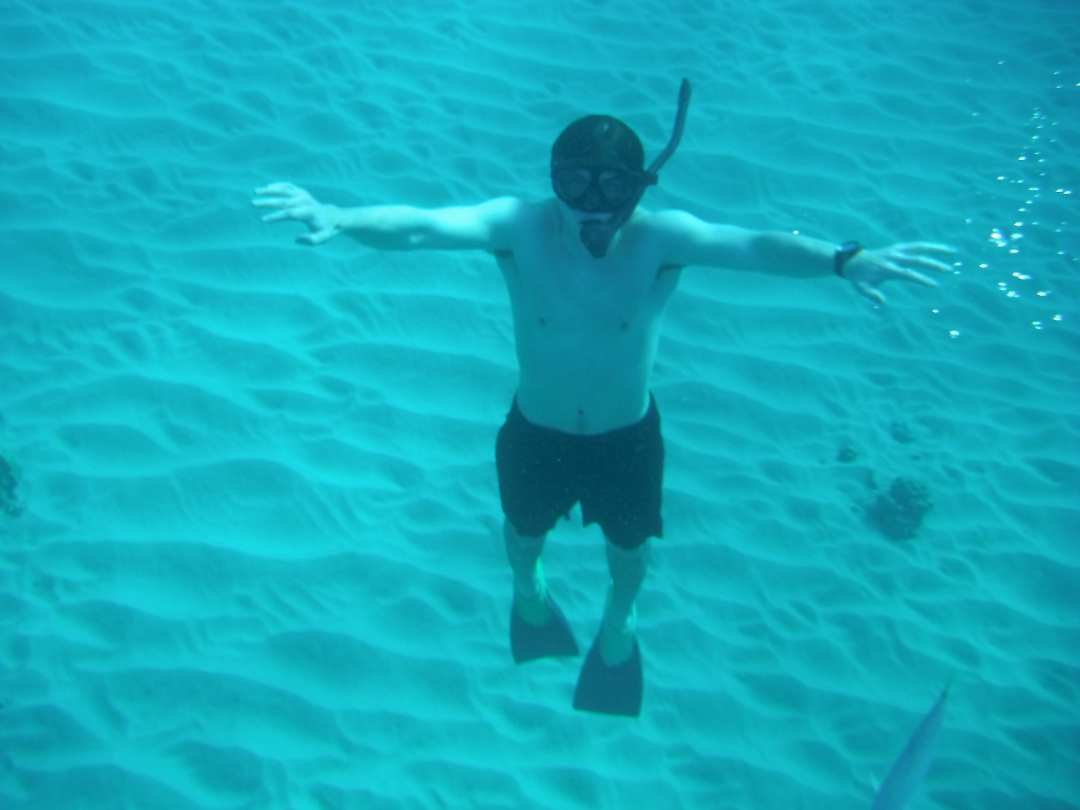}
        \caption{}        
    \end{subfigure}
    \caption{Example non-standard diver poses that are typical during scuba diving operations. Diver robot interaction scenarios must accommodate these poses to be useful for underwater missions.}
    \label{fig:nonstandard_pose}
\vspace{-6mm}
\end{figure}

Many techniques have been devised to establish both R2H~\cite{fulton2022hreyes} and H2R communication (\eg\cite{enan2022robotic}). 
These systems are supported by complementary techniques to enable visual robot control to place the robot within a safe distance of the human~\cite{fulton2022using}. 
This allows higher fidelity understanding from both the robot's and the human's perspective, since it is thought that the information exchange is best when interpreted from the alignment between the human's eyes and the robot's camera~\cite{fulton2022using}. To ensure robust communication, authors in~\cite{enan2022visual} used a transformer-driven network for detecting diver gaze based on facial mask keypoints. However, their work does not handle general poses Fig.~\ref{fig:nonstandard_pose}, or those in which the facial keypoints are not visible.  
 
We argue that a complementary problem to the works of~\cite{fulton2022using} and~\cite{enan2022visual} is utilizing a two-dimensional pose estimator, along with a stereo visual approach to establish three-dimensional positions. By doing this, we can accommodate non-standard human poses, such as those shown in Fig.~\ref{fig:nonstandard_pose}. This will ultimately enable more complex robotic control for re-orientation; \eg when the human is conducting complex tasks and is unable to re-orient themselves with respect to the robot, the robot can come to the human. It is from this perspective that we define our primary contributions to support UHRI, which can be summarized as follows:
\begin{itemize}
    \item The aggregation of a diverse torso keypoint dataset and results from training an off-the-shelf pose estimation algorithm for two-dimensional human pose estimates that accommodates non-standard body poses.
    \item The computation of an alignment vector and establishment of a convention for assigning a coordinate frame to a human's facing direction.
    \item Scale preserving setpoint computations which preserve different body shapes at different distances between the robot and the human.
\end{itemize}

\section{Related Work}
The work introduced in this paper exists within the boundary between UHRI and computer vision for pose estimation. Here we discuss some of the works that influence our methodology. 

\textbf{Underwater human robot interaction}. Various methods have been proposed for robotic detection and individual identification of human divers (\eg\cite{xia2019visual,islam2017mixed, sattar2007your}) with features extracted from visual, or spatio-temporal signals. Others have utilized sonar detection mechanisms to both directly detect in frequency space the presence of a diver \cite{kvasic2019convolutional} and reconstructed acoustic images \cite{lo2012diver, demarco2013sonar}. For explicit communication between an AUV and divers, both robot-to-human and human-to-robot methods have been proposed; \eg robots have used  light~\cite{fulton2022hreyes}, motion~\cite{fulton2019robot,fulton2022robot}, and other cues to communicate intent and information to divers, and fiducials~\cite{Sattar2007ICRA,Sattar2008ICRA,sagitov2017artag}, hand gestures~\cite{gesturesAUV2011,chiarella2018novel, islam2019understanding,GomezChavez2021RAM}, and complex user interface devices~\cite{Verzijlenberg2010IROS,Bernardi2019Diver} have all been used by divers to control robots. However, it is conceivably challenging and constraining for divers to use tags or UI devices while underwater for certain tasks.

\textbf{Human Pose Estimation}. Pose estimation is the task of determining a set of keypoints that define human joint positions in an image. 
Various techniques exist, but most rely on convolutional neural networks (CNNs)~\cite{goodfellow2016deep} to perform feature extraction and output heatmaps over candidate locations \cite{xiao2018simple, dai2022fasterpose, cao2017realtime}. 

The networks are trained to regress from heatmaps to perform keypoint localization by selecting the location with the highest probability as the most likely joint location.

Localizing joint locations accurately is a significant challenge underwater, which is exacerbated by inconsistent lighting conditions and the lack of saliency, or pronounced features, within the typical diver silhouette. Chavez \etal~utilize a recurrent neural network (RNN) with long short-term (LSTM) cells to learn the sequential joint orientations affixed to the human diver, exploiting stereo vision and $17$ inertial measurement units (IMUs) that communicate the diver's movements acoustically.
We recognize that placing additional burden on the diver's already intense cognitive load is problematic. 
Instead, we endeavor to construct F2F re-orientation in such a way that the robot re-orients itself with respect to the human based off of image observations alone\starnote{first use of F2F}.
\section{The Face-to-Face Reorientation Approach}
\label{sec:f2f_approach}
The F2F scale-preserving setpoint computation comprises two components.
First, a pose estimation component localizes torso keypoints, and second, an alignment vector computation establishes a convention for affixing a right-handed coordinate system to the human, from which we compute the transformation that anti-aligns the body frame coordinate system with the camera frame. 
Perspective projection allows us to recover the ideal setpoint, which is the configuration in which the human is facing the camera. 

\begin{figure}[t]
	\vspace{3mm}
    \centering		\includegraphics[width=1.0\columnwidth]{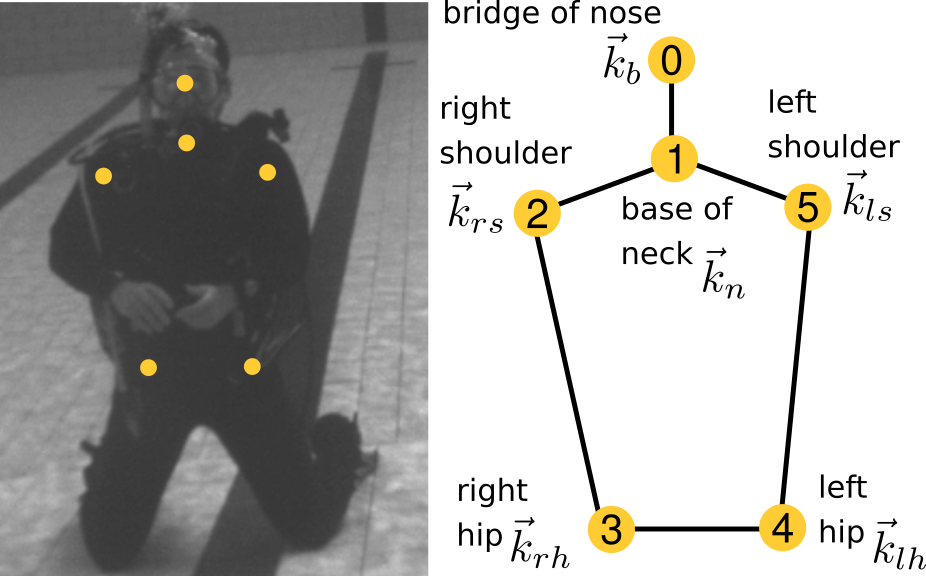}
    \caption{Pose keypoint convention used by F2F along with a sample labeled image. The pose estimator may provide more anatomical keypoints, but F2F only requires these six.}    
    \label{fig:pose_keypoints}    
\end{figure}

\begin{figure}[t]
\vspace{3mm}
\captionsetup[subfigure]{labelformat=empty}
    \centering
    \vspace{2mm}
    \begin{subfigure}[t]{0.32\columnwidth}
        \includegraphics[width=1.0\textwidth]{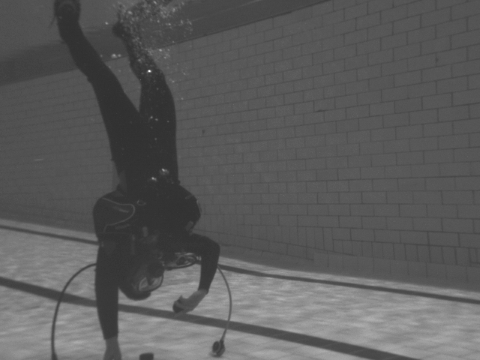}
        \caption{}
    \end{subfigure}
    \begin{subfigure}[t]{0.32\columnwidth}
        \includegraphics[width=1.0\textwidth]{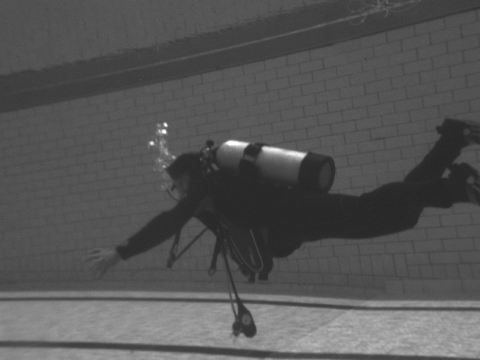}
        \caption{}        
    \end{subfigure}
    \begin{subfigure}[t]{0.32\columnwidth}
        \includegraphics[width=1.0\textwidth]{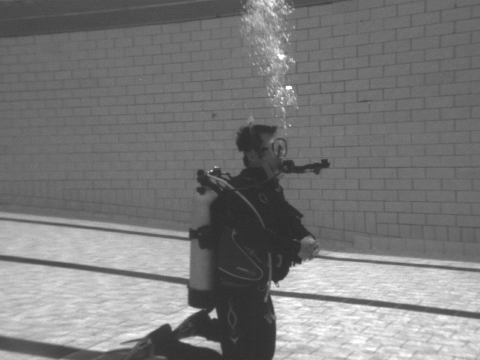}   
        \caption{}        
    \end{subfigure}         
    \caption{Sample raw images from the F2F pose dataset. The dataset contains diverse poses to represent the broadest possible set of orientations a diver can assume while conducting underwater operations.}
    \label{fig:f2f_poses}
    \vspace{-5mm}
\end{figure}

While a future goal is to use a three-dimensional pose estimation algorithm on monocular camera data, much of the work in three-dimensional pose estimation first uses multi-camera setups to triangulate pose keypoints to provide a z-component to ground truth labeled data. 
Pose estimation algorithms can then be trained to directly predict a three-dimensional vector from a single monocular image, see \cite{yu2020humbi, yoon2021humbi}, for example. 
In the underwater domain, instrumenting an experimental setup with calibrated multi-view cameras is a challenge and impractical for most setups. 
To that end, we utilize calibrated stereo cameras affixed to a robot to collect and aggregate images of human divers. 
We trained a deep neural network on labeled image data to localize two-dimensional human joints. 
During runtime, we utilize stereo reconstruction based on pose alone. 
Dense stereo matching is found to be ineffective for the underwater environment. 
The typical diver silhouette almost entirely appears the same to traditional block matching algorithms or even more sophisticated post-processing techniques that fill in gaps in the disparity map. 
From the scale-preserving setpoint computation, we preserve different body shapes to ensure that the robot can automatically predict the optimal alignment or setpoint for the visual control scheme, without need for human intervention or calibration before the beginning of the mission. 
Instead, the robot can detect the setpoint using our method and perform classic visual control schemes. 
A detailed account of visual servo control schemes is beyond the scope of the present work, but an extensive treatment can be found in \cite{chaumette2006visual, chaumette2008visual, chaumette2016visual}.
\textbf{Stereo Diver Pose Dataset}. We aggregated and labeled $6,711$ stereo image pairs from a closed-water environment in which a stereo vision camera was used to collect images of divers in diverse poses. 
The images collected are all of size  $640\times 480$ pixels, and feature a \textit{single} diver in a full wet suit and dive gear. 
The hands and face are exposed, except for parts of the face, which are partially obscured by the breathing regulator and mask. 
The poses were labeled according to the convention shown in Fig.~\ref{fig:pose_keypoints}. 
We argue that the points that encompass the torso, the midpoint of the eyes, and the base of the neck, are the primary keypoints for robotic re-orientation. 
The torso keypoints do not change relative to each other throughout normal diving operations, whereas the limbs are often moving to stabilize the diver's position in the water. 
The midpoint of the eyes was also utilized to accurately reflect the midpoint of the body itself. 
Without this additional keypoint, the robot would come face-to-face lower than expected with the diver, inhibiting the most efficient communication. 
Example poses are shown in Fig.~\ref{fig:f2f_poses}.


\textbf{Pose Estimation}. With the aggregated dataset, we train a deep neural network based on the DeepLabCut (DLC) framework~\cite{mathis2018deeplabcut, nath2019using} for torso keypoint estimation. 
Specifically, we train a ResNet-50-based neural network using $95$\% of the dataset, or approximately $6,375$ images, for $500,000$ training iterations. 
We find the test error to be $12.75$ pixels, and the train error at $11.14$ pixels. 
We then use a threshold \textit{p}-cutoff of $0.05$ to condition the (x,y) coordinates for future analysis. 
While the error might seem high, our goal is to quantify the extent to which we consider alignment as face-to-face interaction. 
To that end, we do not need nearly-optimal keypoint configurations, but localizations that are good enough to recover sparse 3D reconstructions are sufficient; we find that approximately $10$ pixels of error provides enough accuracy for our application.

Examples from the pose estimator evaluated on the test dataset are shown in Fig.~\ref{fig:dlc_output} below. 
Generally, DLC performs adequately localizing the joints. 
However, there are times where even variations in labeling likely contribute to issues in localization. 
For example, if the joint is occluded, it is possible the DLC network will be unable to predict location.
\begin{figure}[h]
\vspace{2mm}
\captionsetup[subfigure]{labelformat=empty}
    \centering
    \begin{subfigure}[t]{0.49\columnwidth}
        \includegraphics[width=1.0\textwidth]{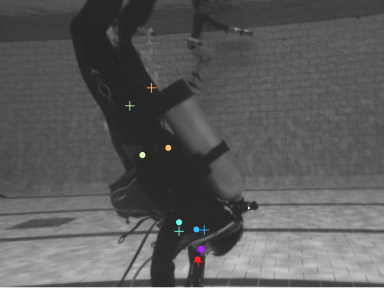}
        \caption{}
    \end{subfigure}
    \begin{subfigure}[t]{0.49\columnwidth}
        \includegraphics[width=1.0\textwidth]{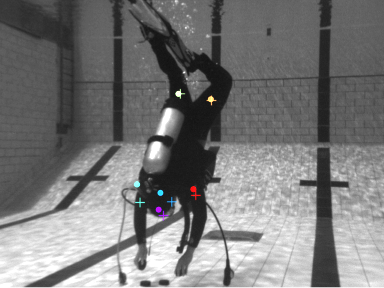}   
        \caption{}        
    \end{subfigure}       
    \caption{DeepLabCut evaluation on the F2F test dataset. Cross markers indicate ground truth labels, and dots indicate DeepLabCut estimations with confidences $p > p_{\text{cutoff}} = 0.05$.}
    \label{fig:dlc_output}
\end{figure}

\textbf{Alignment Coordinate System Convention}. A key contribution of this work is the computation of the alignment vector. The \textit{alignment vector} extends perpendicularly to the plane defined by the torso of the diver. 
This serves as the anchor vector, which allows us to define a coordinate system that is attached to the diver. 
To construct this vector, we use a set of joint location keypoints from our pose estimation network.
Let $\vec{k}_{\text{b}}$ define the bridge of the nose, $\vec{k}_{\text{n}}$ define the base of the neck, $\vec{k}_{\text{rs}}$ define the right shoulder, $\vec{k}_{\text{rh}}$ define the right hip, $\vec{k}_{\text{lh}}$ define the left hip, and $\vec{k}_{\text{ls}}$ define the left shoulder.

After performing stereo image rectification and triangulating based on joint localizations from both the left and right image pairs, the vectors $\vec{k}_i\in\mathcal{K}$, where $i$ corresponds to a specific joint, are defined in three-dimensional camera coordinates as $({}^{c}x_i, {}^{c}y_i, {}^{c}z_i)$. Let $\vec{s}_i$ denote the corresponding projection of the point $\vec{k}_i$ onto the image plane of the robot's camera. The projection $\vec{s}_i = (u_i, v_i)$ is a point defined in the image plane, where $u_i \in [0, N]$ and $v_i \in [0, M]$, and $N,M$ are the image width and height, respectively.
To compute the alignment vector, we define the following steps.
\vspace{-0.25mm}
\change
\begin{enumerate}

\item We compute the center of the predicted keypoints as $\vec{k}_{\text{o}} = \langle\vec{k}\rangle_{\vec{k}\in\mathcal{K}}$, where $\langle\cdot\rangle$ defines the vector average computation.

    The resultant vector $\vec{k}_{\text{o}}$ is located approximately center of mass and is skewed toward the upper part of the torso.

    \item We define several difference vector quantities that exist on the torso plane as \starnote{Can you say why these are needed?}
    \begin{align}
         \vec{k}_{\text{lsh}} &= \vec{k}_{\text{ls}}-\vec{k}_{\text{lh}} &
          \vec{k}_{\text{nlh}} &= \vec{k}_{\text{n}}-\vec{k}_{\text{lh}} \\
          \vec{k}_{\text{nrh}} &= \vec{k}_{\text{n}}-\vec{k}_{\text{rh}} & 
          \vec{k}_{\text{rsh}} &= \vec{k}_{\text{rs}}-\vec{k}_{\text{rh}}.
    \end{align}
    These quantities are needed to establish the relationships between joint locations, effectively defining the torso plane and conditioning the proceeding analysis with respect to the torso plane.
    
    \item We compute the alignment direction by taking the average direction of the cross product between the difference vectors of the torso and neck joints. This defines a direction perpendicular to the plane defined by the torso keypoints
    \begin{align}
         \vec{k}_{\text{l}_{\times}} &= 
         \vec{k}_{\text{lsh}} \times \vec{k}_{\text{nlh}}\\
          \vec{k}_{\text{r}_{\times}} &= \vec{k}_{\text{nrh}} \times  
          \vec{k}_{\text{rsh}}.
    \end{align}    
    \noindent To compute the average direction and define a unit vector, we first take the average, and then we divide by the vector $L2$-norm
    \begin{equation}
        {}^{c}\hat{z}_{B} \equiv \frac{\langle\vec{k}_{\text{r}_{\times}},\vec{k}_{\text{l}_{\times}}\rangle}{\lVert\langle\vec{k}_{\text{r}_{\times}},\vec{k}_{\text{l}_{\times}}\rangle\rVert_2}.
        \label{eqn:alignment_vector}
    \end{equation}
    The alignment vector given by (\ref{eqn:alignment_vector}) points in a direction perpendicular to the plane defined by the torso keypoints. We now affix a right-handed coordinate system to $\vec{k}_{\text{o}}$, with ${}^{c}\hat{z}_{B}$ aligned along the direction given in (\ref{eqn:alignment_vector}). 

   We choose ${}^{c}\hat{y}_{B}$ to be the vector that points along the direction of the midpoint between hip joints. This is given by computing the midpoint of the line segment connecting the hip joints
    \begin{equation}
        \vec{k}_{\text{midpt}} = \langle \vec{k}_{\text{lh}},\vec{k}_{\text{rh}} \rangle.
    \end{equation}
    \item From this we compute the unit vector that points from the center of mass 
 vector $\vec{k}_{o}$ to $\vec{k}_{\text{midpt}}$. This unit vector is defined to be   ${}^{c}\hat{y}_{B}$
    \begin{equation}
         {}^{c}\hat{y}_{B} = \frac{\vec{k}_{\text{midpt}}-\vec{k}_{o}}{\lVert \vec{k}_{\text{midpt}}-\vec{k}_{o} \rVert_2}.
    \end{equation}
    \item Finally, the ${}^{c}\hat{x}_{B}$ is computed through a cross product ${}^{c}\hat{x}_{B} ={}^{c}\hat{y}_{B} \times {}^{c}\hat{z}_{B}$.
    Together these constitute the body frame ${}^{c}\mathcal{F}_{B} = [{}^{c}\hat{x}_{B},{}^{c}\hat{y}_{B},{}^{c}\hat{z}_{B},\vec{k}_{\text{o}}] $ of the human diver, affixed to the midpoint of the extracted pose keypoints, with the ${}^{c}\hat{z}_{B}$ aligned
    in the direction perpendicular to the plane defined by the torso keypoints. It is from this that we can then compute the ideal pose configuration by anti-aligning the body frame with the camera frame unit vectors. 
\end{enumerate}
\stopchange 

The preceding analysis computes the body frame ${}^{c}\mathcal{F}_{B}$ with respect to the camera frame. To compute the pose setpoint for a visual control scheme, we need to compute the transformation that anti-aligns the camera frame and the body frame. That is, there exists some transformation $\tilde{\text{T}}$ that yields an ideal configuration of the body frame ${}^{c}\mathcal{F}^{*}_{B}$
\begin{equation}
    {}^{c}\mathcal{F}^{*}_{B} = \tilde{\text{T}}{}^{c}\mathcal{F}_{B}.
\end{equation}

\noindent This transformation is not known \textit{a priori}.

For a camera that has z-axis along the optical axis, a right-facing x-axis, and a down-facing y-axis,  the following constraints hold 

\begin{align}
    {}^{c}\hat{z}^{*}_{B}\cdot{}^{c}\hat{z} &= -1 & 
    {}^{c}\hat{x}^{*}_{B}\cdot{}^{c}\hat{x} &= -1 &     {}^{c}\hat{y}^{*}_{B}\cdot{}^{c}\hat{y} &= 1.
\end{align}

From these equations, we compute the rotation matrix required to align the body frame with respect to the camera frame to be in an F2F orientation as

\begin{equation}
    {}^{c}\mathcal{F}^{*}_{B} = \tilde{\text{T}}{}^{c}\mathcal{F}_{B}.
\end{equation}

The transformation that aligns the axes can be computed using the Kabsch algorithm \cite{Kabsch:a15629} which minimizes a root-mean-square error function to find the optimal rotation matrix that aligns a set of vectors. The keen observer will note that this alignment does not quite yield the configuration desired. In fact, it aligns the coordinate systems such that the human would be facing away from the robot's camera. To anti-align the coordinate systems, we define $\tilde{\text{T}} \gets \text{R}_y(\pi)\tilde{\text{T}}$. This produces the desired rotation.

Together, along with a translation constraint, which defines how close the keypoints should appear to the camera frame, or the robot, these can be used to compute the components of the transformation matrix. Finally, perspective projection using the camera intrinsics yields the appearance of the points in the image plane by $\vec{s}^{*} = \text{K}\tilde{\text{T}}\vec{k}$. The vector of points $\vec{s}^{*}$ is the scale-preserved setpoint.

\begin{figure}[t]
\captionsetup[subfigure]{labelformat=empty}
    \centering
    \vspace{2mm}
    \begin{subfigure}[t]{0.49\columnwidth}
        \includegraphics[width=1.0\textwidth]{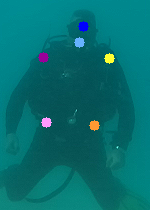}
        \caption{}
    \end{subfigure}
    \begin{subfigure}[t]{0.49\columnwidth}
        \includegraphics[width=1.0\textwidth]{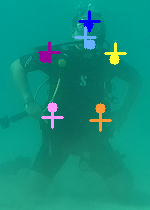}   
        \caption{}        
    \end{subfigure}        
    \caption{Two divers of different body shapes have different setpoints at the same scale, in this case 2 m from the camera. The crosses in the right-hand figure are the baseline setpoints from the diver in the left-hand image. The points have been shifted to center on the mean of the keypoints in the right-hand image. F2F produces scale aware setpoints, such that the robot can utilize visual servo techniques to ensure safe approach from non-standard body poses.}
    \label{fig:divers_scale}
\end{figure}

The benefits of this approach are two-fold. By computing a body-fixed frame, we can compute the setpoint for a visual-servo control scheme that is scale-aware. This means that different diver body shapes will appear as different sizes depending on the scale heuristic. For example, the divers shown in Fig.~\ref{fig:divers_scale} are at the same distance from the camera, but the ideal setpoint is very different. 

As a result, the control scheme would indicate that the robot should move closer to achieve the same level of error between the setpoint and the observed pose. Of course, this could potentially place the diver in harms way if the robot malfunctions. 
\section{Experimental Results}
\label{sec:results}
 Experiments were conducted using data collected from divers in both closed-water (\ie pool) settings and in the ocean waters off the coast of Barbados, West Indies. For this work, we focus on measuring the error between baseline setpoints and constructed setpoints using our alignment vector convention. \change We also provide additional analysis of different pose estimation networks on our custom dataset in the accompanying video submission.\stopchange
\begin{figure}[t]
\captionsetup[subfigure]{labelformat=empty}
    \centering
    \vspace{2mm}
    \begin{subfigure}[t]{0.32\columnwidth}
        \includegraphics[width=1.0\textwidth]{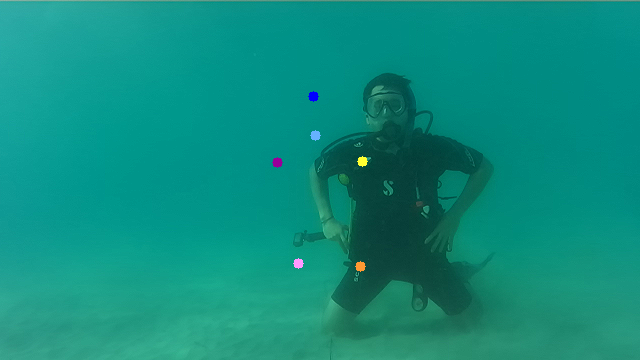}
        \caption{1 m}
    \end{subfigure}
    \begin{subfigure}[t]{0.32\columnwidth}
        \includegraphics[width=1.0\textwidth]{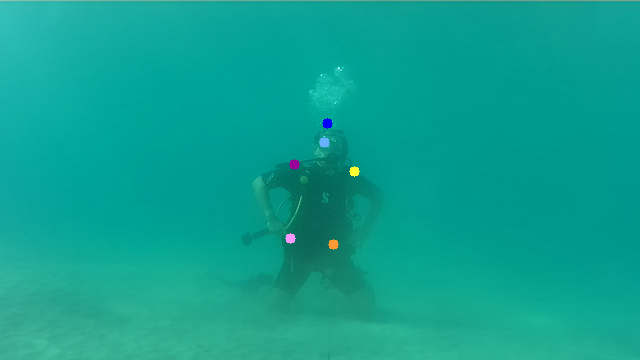}   
        \caption{2 m}        
    \end{subfigure}      
    \begin{subfigure}[t]{0.32\columnwidth}
        \includegraphics[width=1.0\textwidth]{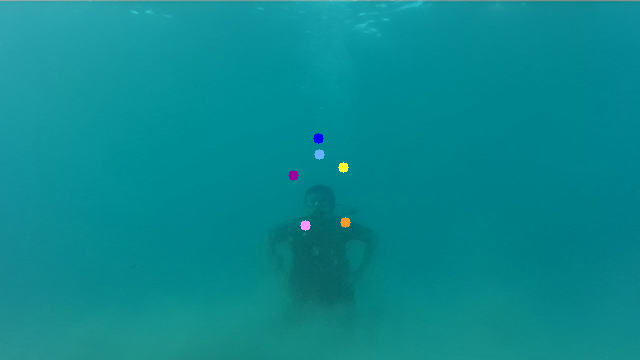}   
        \caption{3 m}        
    \end{subfigure}           
    \caption{Setpoints used as baselines for comparing projections from reconstruction and alignment. Note that setpoint baselines (colored dots) have been shifted to the center of the image to accommodate for differences in image capture that occurred due to strong ocean currents that made station keeping challenging for the divers.}
    \label{fig:divers_setpt_baseline}
\end{figure}
\vspace{20mm}
\begin{table*}[ht]
\renewcommand{\arraystretch}{1.2}
\vspace{2mm}
\caption{Summary of projection errors between setpoint baselines at $1$, $2$, and $3$ m distances and projections from scale-preserving computations, averaged over $50$ projection estimates. Errors are reported as \textit{mean}$\pm$\textit{standard deviation} in pixel units. Each error is the sum over all keypoints of the Euclidean distance between the ground truth and the predicted keypoint location.}
\small
\begin{center}
\newcolumntype{R}{>{\centering\arraybackslash}X}%
\begin{tabularx}{\columnwidth}{l |R|R|R|R }
\toprule
{\textbf{Pose}}&$\boldsymbol{1}~\textbf{m}$&$\boldsymbol{2}~\textbf{m}$&$\boldsymbol{3}~\textbf{m}$&\textbf{Across distances}\\
    \cline{1-5}
    Prone (surface) & $175.39\pm62.67$ & $227.19\pm8.3$ & $322.61\pm0.0$&$241.73\pm23.66$\\ 
Prone (bottom) & $181.43 \pm 64.31$ & $192.04  \pm0.0$ & $157.98\pm 6.51$ & $177.15 \pm 23.6$\\ 
Upright (away) & $367.4 \pm 125.43$ & $61.9 \pm 0.98$ & $88.81\pm 0.0$  & $173.99 \pm 52.7$\\ 
Upright (facing) & $166.84 \pm 0.0$ & $383.12 \pm 157.83$ & $92.67\pm31.70$  & $214.21 \pm 63.17$\\ 
Inverted (facing) & $176.71 \pm 43.63$ & $101.43 \pm 12.77$ & $224.08 \pm 71.63$ & $167.41 \pm 42.68$ \\ 
Inverted (away) & $366.11 \pm 36.35$ & $172.28 \pm 43.43$ & $131.31 \pm 58.75$ & $223.23 \pm 46.18$\\ 
\cline{1-5}
\textbf{Across poses} & $238.98 \pm55.40$ & $189.66 \pm37.22$ & $170.22 \pm33.38$\\     
\bottomrule
\end{tabularx}
\end{center}
\label{table:error}
\end{table*}
\begin{figure*}[t]
\captionsetup[subfigure]{labelformat=empty}
    \centering
    \begin{subfigure}[t]{0.32\columnwidth}
        \includegraphics[width=1.0\textwidth]{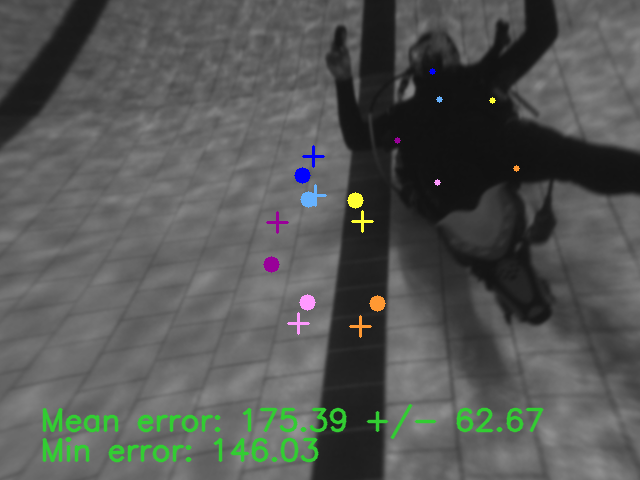}
        \caption{Prone (surface)}
    \end{subfigure}
    \vspace{2mm}
    \begin{subfigure}[t]{0.32\columnwidth}
        \includegraphics[width=1.0\textwidth]{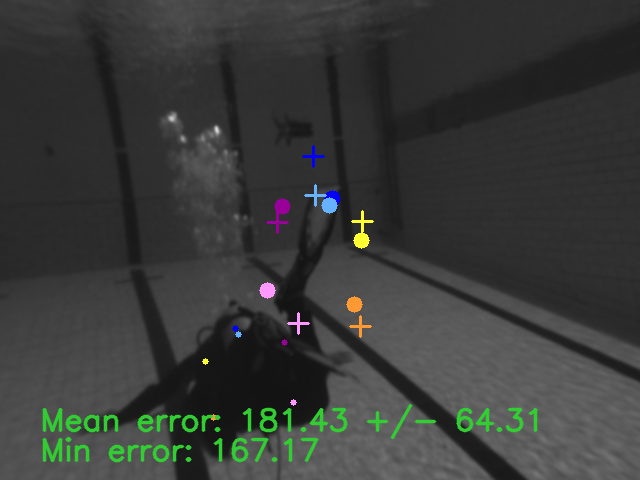}   
        \caption{Prone (bottom)}        
    \end{subfigure}      
    \begin{subfigure}[t]{0.32\columnwidth}
        \includegraphics[width=1.0\textwidth]{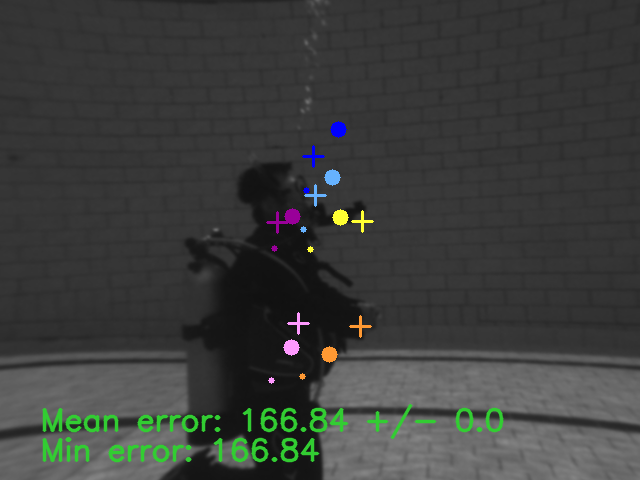}   
        \caption{Upright (facing)}         
    \end{subfigure}   
    \begin{subfigure}[t]{0.32\columnwidth}
        \includegraphics[width=1.0\textwidth]{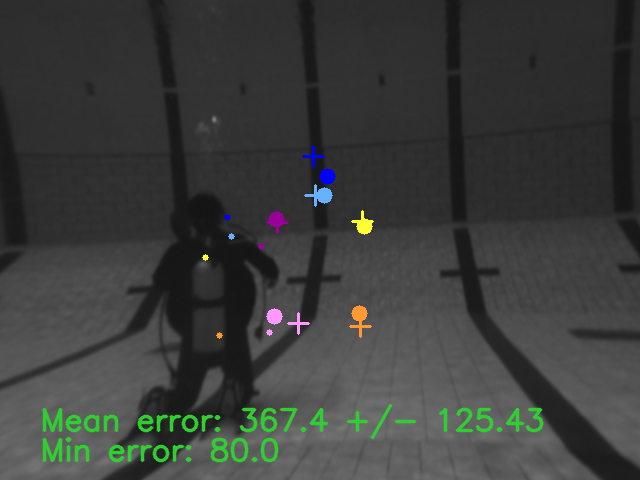}
        \caption{Upright (away)}
    \end{subfigure}
    \begin{subfigure}[t]{0.32\columnwidth}
        \includegraphics[width=1.0\textwidth]{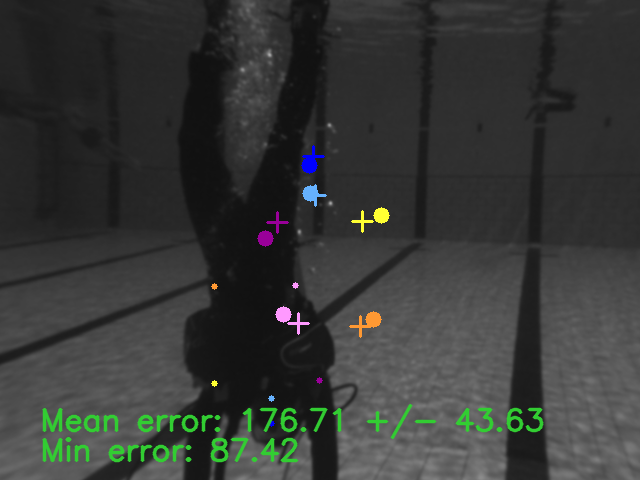}   
        \caption{Inverted (facing)}        
    \end{subfigure}      
    \begin{subfigure}[t]{0.32\columnwidth}
        \includegraphics[width=1.0\textwidth]{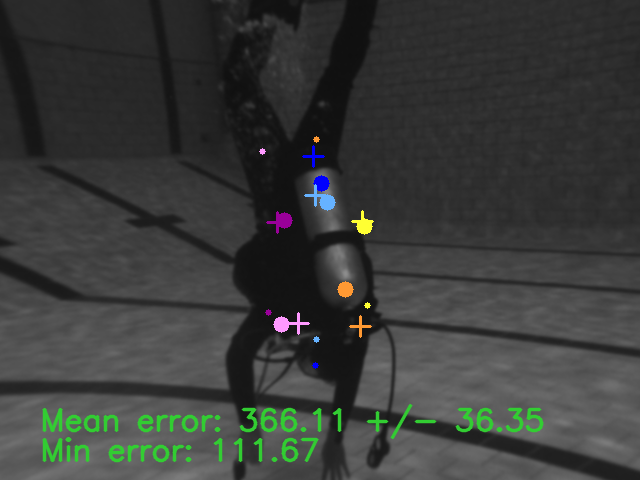}   
        \caption{Inverted (away)}         
    \end{subfigure}      
    \caption{The projections we achieve through the alignment system are reasonable, given the challenges of reconstruction. The cross markers indicate the $1$~m setpoint baseline computed from sea trial data. The large dots indicate projections from the alignment system. A standard deviation of $0.0$ indicates instances in which the system was unable to triangulate accept for a single instance during the frame acquisition.}
    \label{fig:projections_with_setpoints}
\end{figure*}

\vspace{-10mm}
To collect setpoint baselines, a diver for whom we had existing pose data was asked to station keep above an experimental trackline. The trackline was measured to $1$, $2$, and $3$~m distances. The camera operator asked the diver to spend approximately $15$ seconds station keeping during acquisition of stereo image data. The camera operator then signaled the diver to move forward. This process was repeated until the diver had been recorded at all three distances. The setpoint baseline was computed by using a simple by-hand label technique that mimicked what a field operator or end-user would do during calibration of a visual servo system. This hand measurement is shown at the three distances in Fig.~\ref{fig:divers_setpt_baseline}.

We then performed experiments using the setpoints extracted from the three distances and comparing the euclidean error and standard deviation of the error between six canonical pose positions. These results are summarized in the table below. Note that $50$ frames of data at each canonical pose were used during the error computation. A reasonable expectation might be on the order of the error of the pose estimation network. In this case, approximately $60$ pixels of error across all projected keypoints is reasonable. Clearly we do not observe that low of error, except in exceptional cases. There are a couple of reasons for this. The pose estimation network runs inference on both the left- and right-hand image pairs, and any fluctuations caused from camera motion or inconsistent lighting conditions can cause significant errors during triangulation. To that end, notice that we have also shown not only the average error in the figures of Fig.~\ref{fig:projections_with_setpoints} but also the minimum error observed during the frame acquisition. Some poses appear exceptionally good qualitatively and tend to agree with the baselines. There is a trend that indicates reconstruction and projection is better at distances that exceed $1$ m. This is likely caused by better pose estimations. Most of the dataset on which we trained DLC was collected at distances of approximately $2$--$3$ m from the diver. As a result, predictions from the pose network are better and more consistent at these distances, likely causing reductions in reconstruction errors throughout the frames used for analysis.

The results summarized in Table \ref{table:error} demonstrate that we achieve reasonable projection errors for most poses. The visual results for the $1$~m poses are shown in Figure~\ref{fig:projections_with_setpoints} for the six canonical pose states.
\section{Future Work}
 
 Scale-preservation and automatic setpoint computation allows a robot to autonomously detect the desired keypoint configuration, without need for prior calibration. This means in a complex mission, the robot has the ability to re-orient itself with respect to the diver to establish F2F communication. However, the work described has shortcomings. First, totally occluded joints pose a problem for reconstruction, because lateral views appear at the same distance. To subvert this, anthropometric data ratios (ADRs), which describe ratios between human limbs, can be used to regularize depth estimates so that totally lateral views do not appear to have joints at the same distance. Another issue with the present approach is that the algorithm does not account for instances where the diver's body is positioned behind an obstacle relative to the camera's origin. Future work will include context-aware approach strategies that utilize the entirety of the scene for understanding the diver's behavior and selecting the best place to approach from, particularly with regard to complex mission tasks. We will conduct control experiments using the F2F system in pool environments, where evaluation of the control can be better constrained. 
The ocean environment is susceptible to external perturbations from currents making it challenging to evaluate the efficacy of the F2F system. 
\vspace{-2mm}
\section{Conclusion}
We have demonstrated that our methodology enables automatic setpoint computation from human diver pose observations, which could permit more complex UHRI scenarios where the robot must infer automatically the best orientation of target features to navigate within a safe distance of the human diver. We have shown that our method works well for $1$, $2$, and $3$~m baselines, except in cases of poses in which a joint is totally occluded and the triangulation is unable to resolve the depth at which the occluded joint appears. 

\bibliographystyle{ieeetr}
\bibliography{bibliography,allbibs,sattar_j}
\end{document}